\pdfoutput=1
\documentclass[runningheads]{llncs}

\usepackage[T1]{fontenc}
\usepackage{graphicx}
\usepackage{amsmath}
\usepackage{amsfonts}
\usepackage{bbm}
\usepackage{hyperref}
\usepackage{xcolor}
\usepackage{colortbl}
\usepackage{multirow}
\usepackage{tabularx}
\usepackage{caption}
\usepackage{subcaption}
\usepackage{cleveref}

\definecolor{tabfirst}{rgb}{1, 0.7, 0.7}
\definecolor{tabsecond}{rgb}{1, 0.85, 0.7}
\definecolor{tabthird}{rgb}{1, 1, 0.7}

\newcommand{\etal}{et~al\mbox{.}}

\begin{document}

\title{Hybrid Analytical–Learned Representations for Preserving Machine-Readable Patterns in Gaussian Splatting}
\titlerunning{Fiducial Marker Splatting}

\author{Diram Tabaa \and
Gianni A. Di Caro}
\authorrunning{D. Tabaa and G. Di Caro}

\institute{Carnegie Mellon University, Doha, Qatar\\
\email{\{dtabaa, gdicaro\}@andrew.cmu.edu}}

\maketitle
\thispagestyle{titlepage}

\begin{abstract}
3D Gaussian Splatting (3DGS) represents scenes as collections of anisotropic Gaussian primitives optimized end-to-end via differentiable rasterization, achieving photorealistic quality with real-time rendering performance. However, the inherently smooth nature of Gaussian kernels makes them poorly suited to high-frequency binary structures such as fiducial markers and QR codes, which standard gradient-based optimization consistently blurs beyond machine readability. We propose a hybrid analytical–learned reconstruction framework that addresses this fundamental limitation of Gaussian-based scene representations. Our method detects fiducial markers in the input imagery and replaces their learned primitives with analytically constructed Gaussians derived from a minimum rectilinear partition and exponential tiling scheme, while the remainder of the scene retains its standard learned representation. A lightweight appearance transfer step adapts marker colours to scene illumination using k-means clustering of the surrounding learned Gaussians, avoiding further iterative optimization. Experiments on real-world indoor and outdoor scenes show that our approach dramatically improves marker decodability, measured by a novel cosine-weighted readability metric, over standard 3DGS, 2DGS, and texture-augmented baselines, while maintaining competitive image quality and reducing primitive count. These results demonstrate a practical path toward simulation-ready scene reconstructions that preserve the functional semantics required by downstream robotic localization, autonomous navigation, and warehouse logistics systems.

\keywords{Gaussian Splatting  \and Fiducial Markers \and 3D Reconstruction.}
\end{abstract}

\section{Introduction}
\label{sec:intro}

3D Gaussian Splatting (3DGS)~\cite{kerbl3Dgaussians} has emerged as a powerful
paradigm for photorealistic novel view synthesis, enabling real-time rendering of
complex scenes with high visual fidelity. Increasingly, reconstructed 3DGS scenes are
employed as high-fidelity simulation environments for training and evaluating robotic
and AI agents~\cite{chenSplatNavSafeRealTime2025a,zhou2024drivinggaussian}, in domains
ranging from warehouse logistics to precision agriculture. For such environments to
serve as faithful proxies of the real world, the reconstruction must preserve not only
the visual appearance of the scene but also the functional integrity of every element
that an agent may rely upon.

For most scene elements, the residual noise inherent in the reconstruction process does
not critically impair downstream utility. However, a class of high-frequency planar
structures exhibits a fundamentally different sensitivity: their utility is governed by
a binary quality-of-service criterion, functioning either correctly or failing entirely.
Fiducial markers, including AprilTags~\cite{olsonAprilTagRobustFlexible2011}, ArUCo
markers, and QR codes, are a prominent example; others include calibration targets and
industrial barcodes. In robotic systems, fiducial markers frequently serve as the
backbone of localization and control pipelines, so a simulation that cannot preserve
their readability loses its value as a training proxy. These markers encode information
in high-contrast binary patterns whose correct decoding demands pixel-accurate
reconstruction of sharp transitions across a dense grid of cells; even slight softening
of cell boundaries can render a marker entirely undetectable by standard decoders. In
this work, we focus on fiducial markers as a representative and practically important
case.

Faithful reconstruction of such patterns is particularly challenging for Gaussian-based
representations. Gaussian kernels are inherently low-pass
filters~\cite{Hamdi_2024_CVPR}, fundamentally ill-suited to the high-frequency step
functions that define binary cells. Recent methods have sought to improve texture
sharpness in 3DGS through per-primitive texturing~\cite{Svitov_2025_ICCV} or
Fourier-space regularization~\cite{zhangFreGS3DGaussian2024}, yet they do not resolve
the core difficulty: all splatting pipelines minimize continuous photometric losses
(e.g., L1, SSIM) that tolerate soft edges and do not optimize for machine readability. A
rendered marker may achieve high PSNR yet remain undecodable if cell boundaries are
blurred by even one or two pixels. This problem is compounded by camera pose
inaccuracies from structure-from-motion and, in many practical deployments, by the fact
that markers are observed from only a limited number of viewpoints.

To address these challenges, we propose a framework
for faithfully reconstructing fiducial markers within Gaussian splatting scenes under
minimal observation conditions, with an approach that generalizes in principle to other
high-frequency planar structures. Unlike existing methods that rely on differential
rendering to iteratively recover marker appearance, our method analytically reconstructs
marker geometry and appearance, bypassing photometric optimization entirely for the
marker regions. This enables accurate reconstruction of even sparsely observed markers,
provided they are readable in at least one input image. We demonstrate that our approach
consistently outperforms several baseline splatting techniques in marker readability, as
measured by decoder success rate, while incurring negligible overhead in computation and
storage. To facilitate future research, we release a
curated evaluation dataset of real-world scenes containing fiducial markers under
diverse viewing conditions.

In summary, our contributions are: (1)~Fiducial Marker Splatting, the first framework
that analytically reconstructs fiducial markers within 3DGS scenes independently of
photometric optimization, requiring only that the marker be decodable in at least one
input image; (2)~significant improvements in marker readability over standard and
frequency-aware baselines, with minimal additional cost, validated in benchmark
scenes; and (3)~a curated evaluation
dataset to support future work in this direction.

\section{Related Work}
\label{sec:related_work}

We review related work in photorealistic scene reconstruction, focusing on methods for recovering high-frequency detail, and highlight the limitations that motivate our approach.

\vspace{-1em}\paragraph{Neural and Gaussian Radiance Fields.}
Neural Radiance Fields (NeRF)~\cite{mildenhallNeRFRepresentingScenes2020} introduced continuous volumetric scene representations optimized via differentiable ray marching, achieving photorealistic novel view synthesis at the cost of lengthy training and slow rendering. Subsequent works improved quality through anti-aliased multiscale reasoning~\cite{barron2021mipnerf,barron2022mipnerf360}, accelerated training with hash-based encodings~\cite{mueller2022instant}, and eliminated neural networks entirely via sparse voxel grids~\cite{fridovich-keilPlenoxelsRadianceFields2022} or tensorial decompositions~\cite{Chen2022ECCV}. 3D Gaussian Splatting (3DGS)~\cite{kerbl3Dgaussians} culminated this trajectory by representing scenes as collections of anisotropic Gaussians, each parameterized by position, covariance, opacity, and spherical harmonics coefficients, and rendered via differentiable rasterization. Its photorealistic quality, competitive with state-of-the-art implicit methods, combined with real-time frame rates and an explicit, editable scene representation, has driven rapid adoption in robotics~\cite{chenSplatNavSafeRealTime2025a}, autonomous driving~\cite{zhou2024drivinggaussian}, and SLAM~\cite{Matsuki:Murai:etal:CVPR2024}. However, although spherical harmonics enable view-dependent appearance, each Gaussian still resolves to a single color per viewpoint, making the representation inherently smooth and  limited in its ability to encode high-frequency binary patterns.\vspace{-1em}
\paragraph{Frequency-Aware and Per-Primitive Texturing Methods.}
Several works address the limited frequency response of Gaussian primitives. FreGS~\cite{zhangFreGS3DGaussian2024} introduces Fourier-space regularization with progressive frequency annealing, encouraging recovery of increasingly fine detail during optimization. GES~\cite{Hamdi_2024_CVPR} generalizes the Gaussian kernel to an exponential family with a learnable shape parameter~$\beta$; when $\beta > 2$, the resulting falloff is sharper than Gaussian, enabling harder edges with fewer primitives. BBSplat~\cite{Svitov_2025_ICCV} takes an orthogonal approach by augmenting 2D Gaussian primitives with learnable per-splat texture and alpha maps, enabling sub-Gaussian spatial variation within a single primitive. While these methods improve general texture sharpness, they remain bound to continuous photometric optimization that tolerates soft edges and does not enforce the binary decodability required by fiducial markers.

\vspace{-1em}\paragraph{Structured Pattern Refinement in Gaussian Splatting.}
The work most closely related to ours is STRinGS~\cite{STRinGS_2026_WACV}, a text-aware selective refinement framework that improves text readability in 3DGS by isolating text regions, performing targeted densification, and applying region-specific learning strategies. STRinGS identifies an analogous problem: structured visual patterns require special treatment because small reconstruction errors imperceptible in photometric metrics can cause complete semantic failure. However, STRinGS fundamentally remains an optimization-based approach that relies on photometric losses to  recover text appearance, with no guarantee that the result will be decodable. In contrast, fiducial markers possess a  standardized binary structure known a priori. Our method exploits this to \emph{analytically construct} marker Gaussians, bypassing photometric optimization entirely and guaranteeing decodability whenever the marker is readable in at least one input image.

\section{Preliminaries}
\label{sec:preliminaries}

We briefly review 3D Gaussian Splatting, which provides the underlying scene representation, and the theory of minimum rectilinear partitions, which informs our pattern encoding strategy. These form the foundation of our method.

\paragraph{3D Gaussian Splatting.}
3DGS~\cite{kerbl3Dgaussians} represents a scene as a collection of anisotropic Gaussian primitives, each parameterized by a mean $\mu_i\!\in\!\mathbb{R}^3$, a covariance $\Sigma_i\!\in\!\mathbb{S}^3_{++}$, an opacity $\alpha_i\!\in\![0,1]$, and view-dependent color encoded via spherical harmonics.
Rendering proceeds by splatting the 3D Gaussians onto the image plane via an $\alpha$-blending formulation:
\begin{equation}
  C(\mathbf{p}) = \sum_{i \in \mathcal{N}} c_i\,\alpha_i\,
  G^{2\mathrm{D}}_i(\mathbf{p})
  \prod_{j=1}^{i-1}\bigl(1 - \alpha_j\, G^{2\mathrm{D}}_j(\mathbf{p})\bigr),
\end{equation}
where $G^{2\mathrm{D}}_i$ denotes the projected 2D Gaussian evaluated at pixel $\mathbf{p}$, and $\mathcal{N}$ is the set of Gaussians sorted by depth.
Scene parameters are optimized end-to-end by minimizing a photometric loss combining $\mathcal{L}_1$ and D-SSIM between rendered and observed images.

Because Gaussian kernels are inherently smooth, they act as low-pass filters in the spatial domain~\cite{guedonSuGaRSurfaceAlignedGaussian2024}.
While this property makes them ill-suited to represent sharp edges when optimized purely through photometric losses, we show in Sec.~\ref{sec:marker_reconstruction} that Gaussians can in fact closely approximate axis-aligned rectangles---and thus the step functions defining marker cells---through a recursive placement scheme that concentrates progressively smaller primitives along cell boundaries.

\paragraph{Minimum rectilinear partitions.}
\label{subsec:minrect}

A \emph{rectilinear polygon} (or \emph{digital blob}~\cite{ferrariMinimalRectangularPartitions1984}) is a finite, four-connected region on a rectangular grid whose boundary edges are axis-aligned.
A \emph{rectangular partition} of such a polygon is a decomposition into non-overlapping, axis-aligned rectangles that exactly cover it.
Ferrari~\etal~\cite{ferrariMinimalRectangularPartitions1984} showed that the minimum number of rectangles in any such partition is determined by the concavity structure of the polygon's boundary.
Concretely, let $N$ be the number of concave vertices (i.e., boundary vertices with interior angle $270^\circ$) and let $L$ be the maximum number of pairwise non-intersecting axis-parallel chords that can be drawn between pairs of concave vertices sharing a row or column (termed \emph{cogrid} vertices).

Then the minimum-order partition has exactly
\begin{equation}
\label{eq:minpart}
  M^{\star} = N - L + 1
\end{equation}
rectangles for a simply connected polygon, and $M^{\star}\!=\!N\!-\!L\!+\!1\!-\!D$ when $D$ holes are present.
Computing $L$ is equivalent to finding a maximum independent set in a bipartite graph whose vertices represent cogrid chords and whose edges represent intersections; this can be solved in $O(N^{2.5})$ via maximum bipartite matching and K\"onig's theorem.

\section{Fiducial Marker Splatting}

\begin{figure}[ht]
    \centering
    \vspace{-3pt}
    \includegraphics[width=0.80\linewidth]{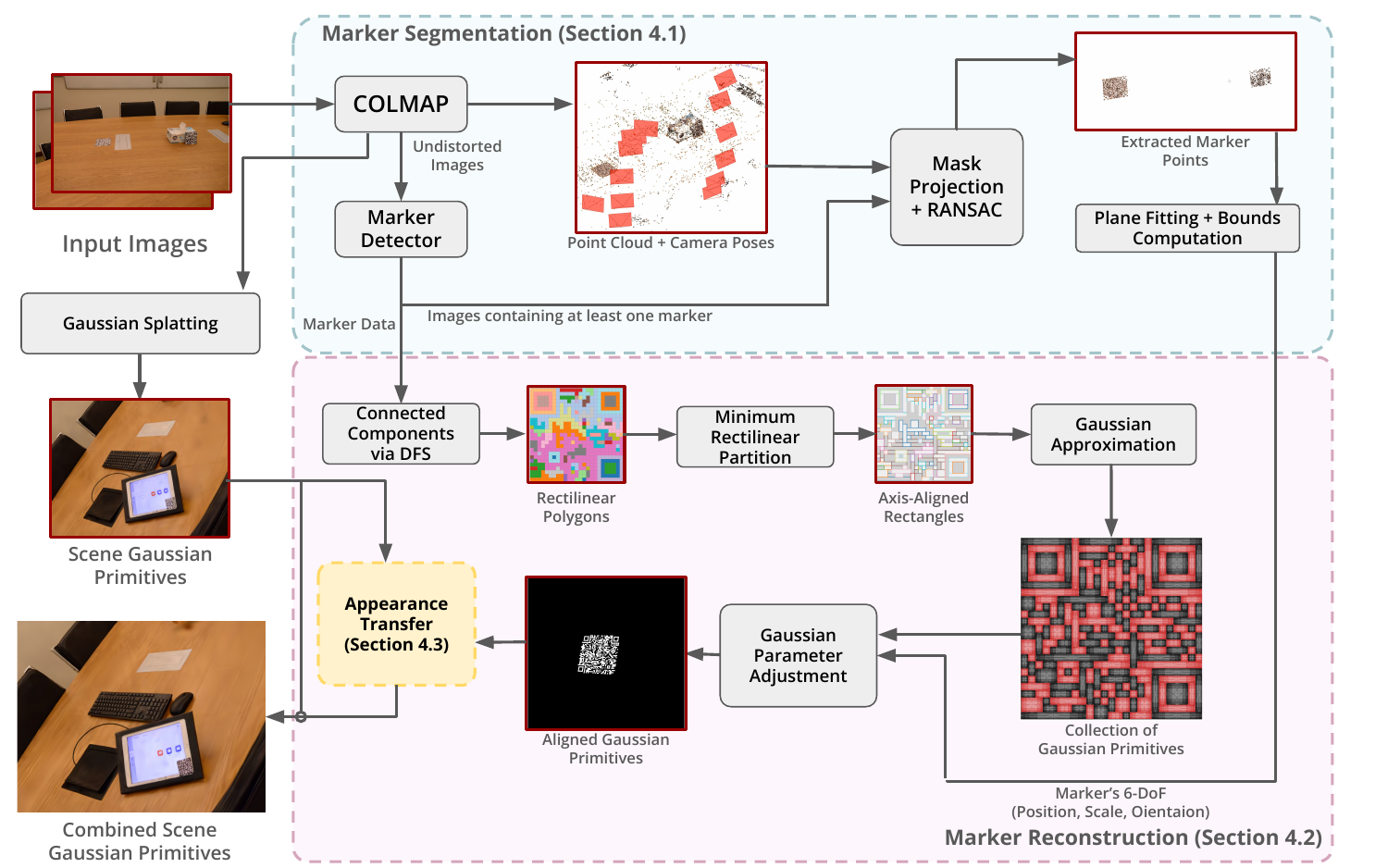}%
    \caption{Overview of our pipeline. Marker segmentation (Sec.~\ref{sec:marker_segmentation}) extracts
marker points from the COLMAP reconstruction via mask projection and RANSAC.
Marker reconstruction (Sec.~\ref{sec:marker_reconstruction}) decomposes extracted regions into
axis-aligned rectangles approximated as Gaussian primitives. Appearance
transfer (Sec.~\ref{sec:appearance}) combines the result with scene Gaussians.}
    \label{fig:fid}
\end{figure}

Replacing a marker inside a trained radiance field is difficult if one relies on
photometric optimisation: the high-frequency, binary nature of fiducial patterns
is poorly served by smooth loss landscapes, and the result is typically a
blurred marker that can no longer be decoded.  Our key insight is that this
optimisation is unnecessary.  Because every standard fiducial marker is a planar grid of axis-aligned monochrome modules, its
Gaussian representation can be written analytically once its pose and payload
are known.  We first recover pose and scale by fusing 2D detections with the
SfM point cloud (\ref{sec:marker_segmentation}), then construct the marker's
Gaussians in closed form (\ref{sec:marker_reconstruction}), and finally match
their appearance to the scene lighting (\ref{sec:appearance}). \Cref{fig:fid} demonstrates our overall pipeline.

\subsection{Marker Segmentation}
\label{sec:marker_segmentation}

The goal of this stage is to recover the 6-DoF pose and metric size of every
fiducial marker visible in the input images.  The key challenge is that a 2D
detector can localise a marker in each frame independently, but cannot directly
provide its 3D geometry.  We bridge this gap by projecting the SfM point cloud
into each detection mask, collecting the 3D points that fall on the marker
surface, and fitting a plane to obtain orientation and scale.  Because the same
marker may appear in several views, we finally merge per-image detections into
a single, more robust estimate per physical marker.

Given a set of posed images, we first obtain a sparse 3D point cloud via
Structure-from-Motion (SfM)~\cite{schoenberger2016sfm}. In parallel, we run a
2D marker detector (e.g., ZBar for QR codes) on each image independently,
yielding a binary mask $\mathcal{M}_i \subset \mathbb{R}^2$ for every detected
marker instance in image $I_i$. Not all images will contain a visible marker;
we denote the subset of images with at least one detection as
$\mathcal{I}_\text{det} \subseteq \{I_1, \dots, I_N\}$. For each image $I_i \in \mathcal{I}_\text{det}$, we project the SfM point
cloud into the image plane using the known camera intrinsics and extrinsics. A
3D point $\mathbf{P}_j$ is \emph{associated} with a marker if its projection
$\pi_i(\mathbf{P}_j)$ falls within the detected mask $\mathcal{M}_i$. Because
the mask is a 2D region, this selection is performed along the full view
frustum and may therefore include spurious points that lie behind the marker
surface. We collect all selected points into a candidate set
$\mathcal{C}_i = \{ \mathbf{P}_j \mid \pi_i(\mathbf{P}_j) \in \mathcal{M}_i \}$.

To separate true marker-surface points from background outliers, we fit a plane
to each candidate set $\mathcal{C}_i$ using RANSAC. Points whose distance to
the consensus plane exceeds a threshold $\tau$ are discarded, leaving an inlier
set $\mathcal{S}_i \subset \mathcal{C}_i$. From $\mathcal{S}_i$ we extract the
marker's surface normal $\mathbf{n}$ via least-squares plane fitting, and
recover the marker's scale from the spatial extent of the inliers along the two
principal in-plane directions. Together, the plane parameters and scale define
the full 6-DoF pose and metric size of each marker instance.

\vspace{-25pt}

Because the same physical marker may be observed from multiple viewpoints, we
merge per-image detections that share a sufficient number of common 3D inlier
points in world space. Specifically, two inlier sets $\mathcal{S}_i$ and
$\mathcal{S}_k$ are consolidated into a single marker instance when
$|\mathcal{S}_i \cap \mathcal{S}_k| / \min(|\mathcal{S}_i|,
|\mathcal{S}_k|) > \alpha$. The merged inlier set yields a more robust
pose estimate, and the decoded marker payload (e.g., the embedded URL or
identifier) is extracted from the highest-confidence detection among the
contributing views. The resulting marker metadata--pose, scale, and decoded
content---is forwarded to the analytic marker generator described in
\ref{sec:marker_reconstruction}.

Our approach assumes that enough SfM points fall on the marker surface to
permit reliable plane estimation. In practice, this assumption holds well:
fiducial markers exhibit high-contrast binary patterns that produce abundant
stable keypoints during feature matching. In cases where the initial SfM
reconstruction is overly sparse, a single round of Gaussian Splatting
densification can be applied to increase point density on the marker surface.
While the densified points may not preserve the marker's visual content
faithfully, they provide sufficient geometric support for accurate surface
normal and scale recovery.

\subsection{Marker Reconstruction}
\label{sec:marker_reconstruction}

The marker segmentation stage (\ref{sec:marker_segmentation}) yields, for each
detected marker, a 6-DoF pose, metric scale, and the decoded payload.  We now
describe how this information is used to synthesize an \emph{analytic} Gaussian
representation of the marker that can be inserted directly into the radiance
field, bypassing any learning-based optimisation of the marker region.

We first decode the marker payload into its canonical binary pattern---an
$M\!\times\!M$ grid of monochrome modules.  To obtain a compact geometric
description, we segment the grid into connected components of identical colour
via depth-first search.  Because each module is axis-aligned and
unit-square, every connected component is a rectilinear polygon.  We then apply
a minimum rectilinear partition (see \cref{subsec:minrect}) to decompose each such
polygon into a minimal set of axis-aligned rectangles.  This minimisation is
important: fewer rectangles translate directly to fewer Gaussian primitives in
the final representation.

Each rectangle in the partition must be represented as a collection of Gaussian
primitives whose superposition closely approximates the rectangle's indicator
function.  A single Gaussian is a poor approximation of a sharp-edged
rectangle; naively increasing the number of uniformly spaced Gaussians is
wasteful.  Instead, we adopt an \emph{exponential tiling} scheme that
concentrates primitives near the rectangle boundaries, where the approximation
error is largest.

Without loss of generality, consider approximating the one-dimensional indicator
$\mathbbm{1}_{[-1,\,1]}(x)$.  We place $2N\!+\!1$ Gaussians indexed by
$k \in \{-N,\dots,0,\dots,N\}$ with centers and scales
\vspace*{-2ex}

\begin{equation}
  \mu_k = \operatorname{sgn}(k)\!\left(1 - 2^{-|k|}\right), \qquad
  \sigma_k = 2^{-|k|}.
  \label{eq:exp_tile_1d}
\end{equation}
At $k\!=\!0$ a single unit-variance Gaussian covers the interior; each
successive level halves the scale and shifts the centre toward the boundary,
forming a telescoping cover that sharpens the transition region with
$\mathcal{O}(\log(1/\epsilon))$ primitives for a boundary approximation error
of~$\epsilon$.

For a two-dimensional rectangle with half-widths $(w_x, w_y)$, the tiling is
constructed as an outer product of two such one-dimensional sequences.
Concretely, let $\{(\mu^x_j, \sigma^x_j)\}$ and $\{(\mu^y_k, \sigma^y_k)\}$
be the 1-D centre--scale pairs obtained from \ref{eq:exp_tile_1d} for widths
$w_x$ and $w_y$ respectively.  Each element of the Cartesian product yields a
2-D Gaussian primitive
\vspace*{-1ex}
\begin{equation}
  G_{jk}(\mathbf{x}) =
    \exp\!\Bigg(
      -\frac{(x - w_x\mu^x_j)^{2}}{2(w_x\sigma^x_j)^{2}}
      -\frac{(y - w_y\mu^y_k)^{2}}{2(w_y\sigma^y_k)^{2}}
    \Bigg),
  \label{eq:exp_tile_2d}
\end{equation}
so that the full rectangle is approximated by
$\mathbbm{1}_R(\mathbf{x}) \approx \sum_{j,k} G_{jk}(\mathbf{x})$.  The
colour attribute of each Gaussian is set to the uniform colour of the
originating connected component, and opacity is set to unit value.

The centres and scales produced by the tiling are defined in a normalised marker
coordinate frame.  We first translate and scale each primitive according to the
rectangle's position within the canonical $M\!\times\!M$ grid, yielding
marker-space coordinates.  The 6-DoF pose recovered in
\ref{sec:marker_segmentation} then provides a similarity transform---rotation,
translation, and metric scaling---that maps these coordinates into the world
frame of the radiance field.  For 3D Gaussian Splatting~\cite{kerbl3Dgaussians},
each primitive is assigned a third scale component $\sigma_z = \varepsilon$
(we use $\varepsilon = 10^{-5}$) to produce effectively planar splats aligned
with the marker surface.  For 2D Gaussian
Splatting~\cite{Huang2DGS2024}, the representation is inherently
surface-based and no out-of-plane scale is required.

\subsection{Appearance Transfer}\label{sec:appearance}

The analytic construction produces geometrically accurate markers with
canonical black and white colours. In practice, scene illumination shifts the
apparent colour of both module types, so we transfer appearance directly from
the pre-trained scene Gaussians rather than running  iterative photometric
optimisation.

After scene training has converged, we extract all scene Gaussians whose means
lie within the 3-D volume of each detected marker, defined by the marker's
pose, metric extent, and a small depth tolerance~$\tau$ along the surface
normal. Because the marker contains exactly two module colours, the zero-order
spherical harmonic coefficients (SH$_0$) of these Gaussians naturally form two
clusters. We apply $k$-means with $k{=}2$ to obtain centroids
$\mathbf{c}_{\text{dark}}$ and $\mathbf{c}_{\text{light}}$, ordered by
$\ell_1$ norm, which capture the scene-specific appearance of the black and
white modules under the prevailing illumination.

Each analytically constructed marker Gaussian then inherits the centroid
corresponding to its module type:
$\text{SH}_0 = \mathbf{c}_{\text{dark}}$ for logical-zero modules and
$\text{SH}_0 = \mathbf{c}_{\text{light}}$ for logical-one modules. All
higher-order SH bands are set to zero to enforce view-independent colour.
The extracted scene Gaussians corresponding to the reconstructed markers are
subsequently removed to avoid visual duplication, and the marker Gaussians are
merged into the scene representation without further training.

\section{Experiments}
\subsection{Settings}

\textbf{Datasets. }
To the best of our knowledge, no existing dataset for novel-view synthesis or
3D reconstruction has been designed with marker readability as an evaluation
criterion. We therefore introduce a new benchmark comprising seven scenes of
varying complexity: three outdoor (\textit{Tomatoes, Sprouts, Rocks}) and four indoor
(\textit{Arcade}, \textit{Library}, \textit{Conference}, \textit{Medicine}). \textit{Tomatoes} and \textit{Sprouts}
depicts a row of plants with two QR codes placed alongside them, simulating a
practical use case in which markers convey information about nearby specimens.
\textit{Arcade} captures an arcade machine bearing a single QR code that
encodes gameplay instructions. \textit{Library} and \textit{Rocks} contains two QR codes embedded among densely textured shelves and rocks, respectively, testing reconstruction under visual clutter.
\textit{Conference} features a meeting room with two QR codes deliberately
placed at different orientations and scales to stress-test the robustness of
the full pipeline.  Lastly, \textit{Medicine} contains both a DataMatrix and a QR code, demonstrating applicability beyond a single marker convention. Across the seven scenes, the dataset spans a range of
lighting conditions, marker sizes, and viewing angles representative of
real-world deployment.

\textbf{Evaluation Metrics.}
Standard image-quality metrics such as PSNR, SSIM, and LPIPS, capture the visual
fidelity of the reconstruction but do not reflect whether a marker remains
machine-readable from novel viewpoints.  We therefore introduce
\emph{Cosine-Weighted Readability} (CWR), a metric that provides a holistic
measure of marker legibility across the full visibility hemisphere while
down-weighting grazing angles, at which successful decoding is inherently more
difficult.  Let $r(\boldsymbol{\omega}) \in \{0,1\}$ denote whether the marker
is successfully decoded from viewing direction $\boldsymbol{\omega}$, and let
$\theta$ be the angle between $\boldsymbol{\omega}$ and the marker surface
normal.  We define
\begin{equation}
  \text{CWR}
  = \frac{1}{\pi}\int_{\Omega^{+}} r(\boldsymbol{\omega})\,
    \cos\theta \;\mathrm{d}\boldsymbol{\omega}\,,
  \label{eq:cwr}
\end{equation}
where $\Omega^{+}$ is the upper hemisphere and the $1/\pi$ factor normalises
the metric to $[0,1]$.  Because evaluating this integral analytically is
intractable, we approximate it via Monte Carlo integration by sampling $N$
directions from a Fibonacci sphere restricted to $\Omega^{+}$, yielding the
self-normalised estimator
\begin{equation}
  \widehat{\text{CWR}}
  = \frac{\sum_{i=1}^{N} r(\boldsymbol{\omega}_i)\,\cos\theta_i}
         {\sum_{i=1}^{N} \cos\theta_i}\,.
  \label{eq:cwr_mc}
\end{equation}

\textbf{Implementation Details.}
Our marker pipeline is implemented in PyTorch, with NumPy and
Shapely used for the geometric decomposition stages.  For both 3D Gaussian
Splatting~\cite{kerbl3Dgaussians} and 2D Gaussian
Splatting~\cite{Huang2DGS2024}, we adopt the gsplat~\cite{ye2025gsplat} rasteriser
with default hyperparameters and train for 16\,000 iterations; we found that
longer training schedules cause the baselines to overfit the training views,
severely degrading CWR on novel viewpoints.  For
BBSplat~\cite{Svitov_2025_ICCV} and GES ~\cite{Hamdi_2024_CVPR}, we use the authors' official implementation and
train for 7\,000 iterations following their recommended configuration.  All
experiments are conducted on a single NVIDIA RTX A6000 GPU.

\subsection{Results}

\textbf{Computational Costs.}
The analytic marker construction introduces negligible overhead relative to standard 3DGS training. The marker segmentation stage (\cref{sec:marker_segmentation})---comprising 2D detection, SfM point projection, RANSAC plane fitting, and multi-view merging---executes in under one second per scene on a single CPU core and is performed only once as a preprocessing step. The minimum rectilinear partition and exponential tiling (\cref{sec:marker_reconstruction}) are likewise computed analytically and add less than two seconds even for scenes containing multiple markers. In total, the marker pipeline adds under five seconds of wall-clock time to the full training procedure, which is dominated by the 16\,000-iteration scene optimisation.

Importantly, the analytic construction also \emph{reduces} the total primitive count in the marker region. Because the minimum rectilinear partition minimises the number of rectangles and each rectangle is tiled with $O(\log(1/\epsilon))$ Gaussians along each axis, the resulting marker representation is substantially more compact than the dense splat clouds that standard densification produces when attempting to fit high-frequency patterns.

\begin{table*}[t]
\begin{small}
    \centering
    \setlength{\tabcolsep}{4pt}
    \renewcommand{\arraystretch}{0.95}
    \caption{{\bf Quantitative Results.}
    (a) Cosine-weighted readability on hemisphere views ($\uparrow$): decoder success rate for each marker across scenes.
    (b) Image quality metrics: PSNR in dB ($\uparrow$), SSIM ($\uparrow$), and LPIPS ($\downarrow$) per scene.
    The best, second, and third results are highlighted in red, orange, and yellow, respectively.}
    \label{tab:cwr}

\begin{minipage}[t]{\linewidth}
        \centering
        \subcaption{Cosine-Weighted Readability ($\uparrow$)}
        \label{tab:hemisphere-readability}
        \resizebox{\linewidth}{!}{
        \begin{tabular}{l|c|cc|cc|cc|cc|cc|cc}
Method
    & Arcade & \multicolumn{2}{c|}{Tomatoes} & \multicolumn{2}{c|}{Library}
    & \multicolumn{2}{c|}{Sprouts} & \multicolumn{2}{c|}{Medicine} & \multicolumn{2}{c|}{Rocks}
    & \multicolumn{2}{c}{Conference} \\
\hline
Ours
    & \cellcolor{tabfirst}0.637
    & \cellcolor{tabfirst}0.881 & \cellcolor{tabfirst}0.919
    & \cellcolor{tabfirst}0.482 & \cellcolor{tabfirst}0.504
    & \cellcolor{tabfirst}0.905 & \cellcolor{tabfirst}0.960
    & \cellcolor{tabfirst}0.691 & \cellcolor{tabfirst}0.067
    & \cellcolor{tabfirst}0.144 & \cellcolor{tabfirst}0.516
    & \cellcolor{tabfirst}0.571 & \cellcolor{tabfirst}0.951\\
3DGS
    & 0.089
    & \cellcolor{tabthird}0.209 & \cellcolor{tabsecond}0.177
    & 0.191 & 0.167
    & \cellcolor{tabthird}0.023 & \cellcolor{tabthird}0.113
    & \cellcolor{tabthird}0.419 & \cellcolor{tabsecond}0.011
    & \cellcolor{tabsecond}0.015 & \cellcolor{tabthird}0.172
    & 0.501 & \cellcolor{tabsecond}0.891\\
3DGS-Antialias
    & 0.023
    & \cellcolor{tabsecond}0.271 & 0.052
    & 0.115 & 0.135
    & 0.000 & 0.030
    & 0.324 & 0.000
    & \cellcolor{tabsecond}0.015 & 0.146
    & 0.519 & \cellcolor{tabthird}0.870\\
2DGS
    & 0.088
    & 0.150 & 0.000
    & 0.178 & \cellcolor{tabthird}0.205
    & \cellcolor{tabsecond}0.076 & \cellcolor{tabsecond}0.264
    & 0.368 & 0.000
    & 0.000 & 0.160
    & \cellcolor{tabthird}0.528 & 0.849\\
BBSplat
    & \cellcolor{tabsecond}0.286
    & 0.015 & \cellcolor{tabthird}0.063
    & \cellcolor{tabsecond}0.290 & \cellcolor{tabsecond}0.296
    & 0.000 & 0.052
    & \cellcolor{tabsecond}0.510 & 0.000
    & 0.000 & \cellcolor{tabsecond}0.188
    & \cellcolor{tabsecond}0.533 & 0.830\\
GES
    & \cellcolor{tabthird}0.141
    & 0.000 & 0.000
    & \cellcolor{tabthird}0.231 & 0.197
    & 0.000 & 0.000
    & 0.386 & 0.000
    & 0.000 & \cellcolor{tabthird}0.172
    & 0.496 & 0.782\\
\end{tabular}
        }
    \end{minipage}

    \vspace{0.4em}

\begin{minipage}[t]{\linewidth}
    \centering
    \subcaption{Image Quality Metrics}
    \label{tab:image-quality}
    \resizebox{\linewidth}{!}{
    \begin{tabular}{l|ccc|ccc|ccc}
\multirow{2}{*}{Scene}
    & \multicolumn{3}{c|}{Ours} & \multicolumn{3}{c|}{2DGS} & \multicolumn{3}{c}{3DGS} \\
    & PSNR & SSIM & LPIPS & PSNR & SSIM & LPIPS & PSNR & SSIM & LPIPS \\
\hline
Arcade
    & \underline{25.81} & 0.896 & \underline{0.324}
    & 25.51 & \textbf{0.904} & 0.337
    & \textbf{26.04} & \underline{0.896} & \textbf{0.324} \\
Tomatoes
    & 26.42 & 0.868 & 0.387
    & \textbf{26.87} & \textbf{0.874} & \textbf{0.383}
    & \underline{26.46} & \underline{0.868} & \underline{0.387} \\
Library
    & 25.95 & 0.934 & 0.336
    & \textbf{26.33} & \textbf{0.944} & \textbf{0.331}
    & \underline{26.16} & \underline{0.937} & \underline{0.334} \\
Sprouts
    & 26.47 & 0.860 & 0.376
    & \textbf{28.19} & \textbf{0.876} & \textbf{0.370}
    & \underline{26.54} & \underline{0.861} & \underline{0.375} \\
Medicine
    & 22.81 & 0.775 & 0.267
    & \textbf{25.39} & \textbf{0.810} & \textbf{0.249}
    & \underline{24.70} & \underline{0.783} & \underline{0.262} \\
Rocks
    & 25.74 & \textbf{0.805} & 0.445
    & \textbf{26.18} & \underline{0.800} & \underline{0.432}
    & \underline{26.09} & 0.797 & \textbf{0.417} \\
Conference
    & 23.94 & 0.852 & \underline{0.417}
    & \textbf{25.20} & \textbf{0.864} & 0.419
    & \underline{24.22} & \underline{0.853} & \textbf{0.415} \\
\hline
Average
    & 25.31 & 0.856 & 0.365
    & \textbf{26.24} & \textbf{0.867} & \underline{0.360}
    & \underline{25.75} & \underline{0.856} & \textbf{0.359} \\
\end{tabular}}
\end{minipage}
\end{small}
\end{table*}

\begin{figure}[h]
\centering
\small
\begin{tabularx}{\linewidth}{*{6}{>{\centering\arraybackslash}X}}
3DGS~\cite{kerbl3Dgaussians} & 2DGS~\cite{Huang2DGS2024} & BBSplat ~\cite{Svitov_2025_ICCV} & \textbf{Ours} \\
\end{tabularx}
\includegraphics[width=\linewidth]{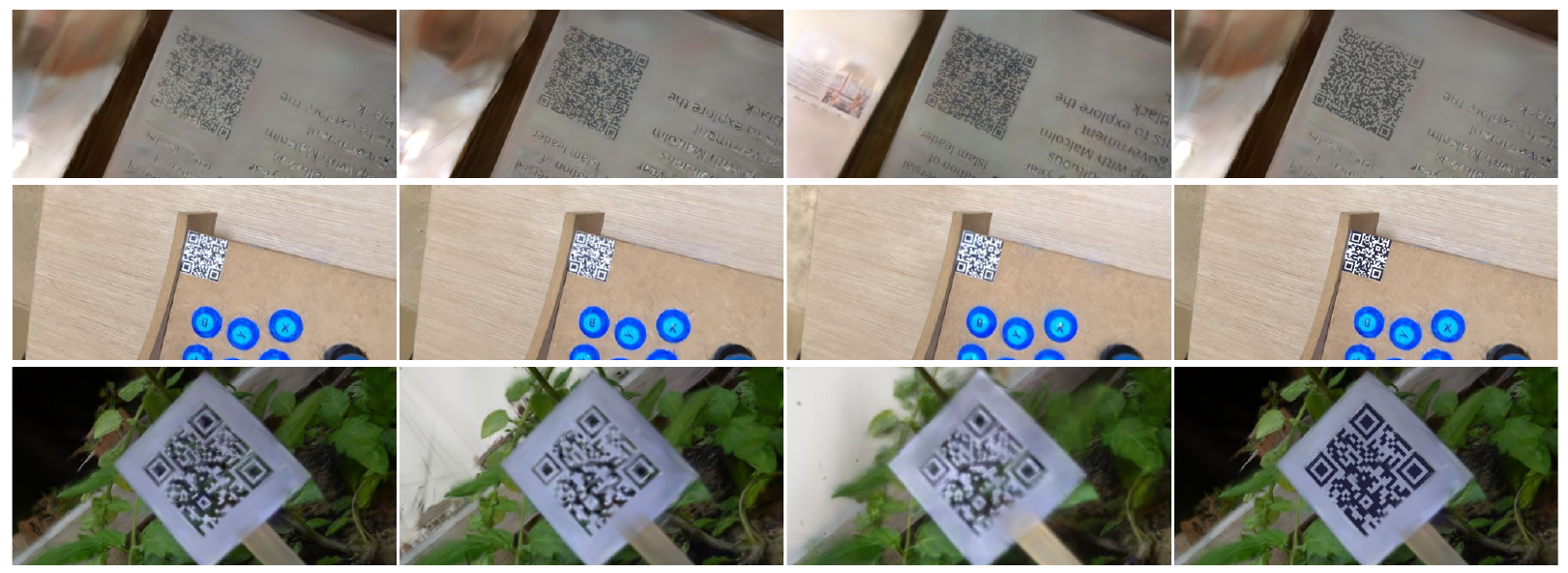}
\captionof{figure}{ Qualitative comparison of QR code reconstruction across methods on the \textit{Library}, \textit{Arcade}, and \textit{Tomatoes} scenes. From left to right: 3DGS ~\cite{kerbl3Dgaussians}, 2DGS ~\cite{Huang2DGS2024}, BBSplat ~\cite{Svitov_2025_ICCV}, and Ours. (Best viewed on screen.)}
\label{fig:qualitative}
\vspace{-3mm}
\end{figure}

\noindent \textbf{Marker Readability.} \quad
Table~\ref{tab:cwr}(a) reports the cosine-weighted readability for each marker across all scenes. Our method achieves the highest CWR on every marker instance, usually by a several-fold margin. The reason is structural: baselines recover markers only as a by-product of photometric fitting, so wherever a marker is small, dark, or finely detailed, the optimisation blurs adjacent cells together and the code becomes undecodable. This is clearest on the smaller QR codes such as Sprouts and Tomatoes, where texture-splat methods like BBSplat collapse almost entirely because their splats cannot resolve the tight black-to-white boundaries, while ours stays above 0.8. The lone hard case for all methods is the Medicine DataMatrix, whose dense, low-redundancy layout is inherently harder to read back than a QR code; even here our method leads, and the residual difficulty reflects the marker family rather than the reconstruction.

The qualitative comparison in Fig.~\ref{fig:qualitative} corroborates these findings: baselines blur individual modules into an indistinguishable gray mass, whereas our method preserves crisp black-to-white transitions that remain machine-readable across viewing angles. This confirms that photometric optimisation alone cannot recover the binary structure of fiducial markers, and that our analytic replacement strategy is essential for reliable decoding.

\noindent \textbf{Photometric Quality.}
Table~\ref{tab:cwr}(b) reports PSNR, SSIM, and LPIPS across all scenes. Crucially, these readability gains come at no meaningful cost to image quality: averaged across scenes our method is on par with both baselines (25.31 / 0.856 / 0.365, versus 25.75 / 0.856 / 0.359 for 3DGS and 26.24 / 0.867 / 0.360 for 2DGS). This is expected, since we replace only the marker regions with analytic Gaussians and leave the rest of the scene untouched, confining any difference to a tiny fraction of each image. The small PSNR gap is most pronounced on Conference (23.94 vs.\ 24.22 for 3DGS), where added markers and varied orientations introduce minor seam artifacts at boundaries. These trade-offs are marginal and imperceptible in practice, and far outweighed by the order-of-magnitude improvement in readability documented above.

\section{Conclusions}

We presented Fiducial Marker Splatting, a framework that analytically
reconstructs fiducial markers within Gaussian splatting scenes, bypassing
photometric optimisation entirely for marker regions. By combining 2D marker
detection with minimum rectilinear partitions and an exponential Gaussian
tiling scheme, our method guarantees marker decodability whenever the marker is
readable in at least one input image. A simple $k$-means colour transfer from
the surrounding scene Gaussians adapts the marker appearance to local
illumination without iterative refinement. Experiments on our  benchmark
of real-world indoor and outdoor scenes show that the proposed approach
substantially outperforms standard 3DGS, 2DGS, BBSplat, and GES in
cosine-weighted readability while maintaining competitive photometric quality
and incurring negligible computational overhead.

\begin{credits}
\subsubsection{\ackname}
This paper was made possible by NPRP grant 14S-0417-210227 from the Qatar Research, Development and Innovation (QRDI) Council / Qatar National Research Fund (QNRF). The statements made herein are solely the responsibility of the authors.
\end{credits}

\bibliographystyle{splncs04}
\bibliography{main}

\end{document}